\documentclass[mlmain]{jmlr}

\usepackage{longtable}
\usepackage{enumitem}

\usepackage{booktabs}
\usepackage[load-configurations=version-1]{siunitx} 

\theorembodyfont{\upshape}
\theoremheaderfont{\scshape}
\theorempostheader{:}
\theoremsep{\newline}

\title[Short Title]{Graph Neural Networks Uncover Geometric Neural Representations in Reinforcement-Based Motor Learning}

\author{\Name{Federico Nardi} \Email{f.nardi21@imperial.ac.uk}\\
\addr Department of Computing, Imperial College London \\ 
UKRI Centre for Doctoral Training in AI for Healthcare, Imperial College London
\AND
\Name{Jinpei Han} \Email{j.han20@imperial.ac.uk}\\
\addr Department of Computing, Imperial College London
\AND
\Name{Shlomi Haar\nametag{\thanks{Equal contribution.}}$^{\dagger}$} \Email{s.haar@imperial.ac.uk}\\
\addr Department of Brain Sciences, Imperial College London \\ UK Dementia Research Institute – Care Research \& Technology, Imperial College London
\AND
\Name{A.Aldo Faisal$^*$\nametag{\thanks{Corresponding authors.}}} \Email{a.faisal@imperial.ac.uk}\\
\addr Department of Computing and Bioengineering, Imperial College London \\ UKRI Centre for Doctoral Training in AI for Healthcare, Imperial College London \\ Chair in Digital Health \& Data Science, University of Bayreuth 
}

\begin{document}

\maketitle

\begin{abstract}
Graph Neural Networks (GNN) can capture the geometric properties of neural representations in EEG data. Here we utilise those to study how reinforcement-based motor learning affects neural activity patterns during motor planning, leveraging the inherent graph structure of EEG channels to capture the spatial relationships in brain activity. By exploiting task-specific symmetries, we define different pretraining strategies that not only improve model performance across all participant groups but also validate the robustness of the geometric representations.
Explainability analysis based on the graph structures reveals consistent group-specific neural signatures that persist across pretraining conditions, suggesting stable geometric structures in the neural representations associated with motor learning and feedback processing. These geometric patterns exhibit partial invariance to certain task space transformations, indicating symmetries that enable generalisation across conditions while maintaining specificity to individual learning strategies.
This work demonstrates how GNNs can uncover the effects of previous outcomes on motor planning, in a complex real-world task, providing insights into the geometric principles governing neural representations. Our experimental design bridges the gap between controlled experiments and ecologically valid scenarios, offering new insights into the organisation of neural representations during naturalistic motor learning, which may open avenues for exploring fundamental principles governing brain activity in complex tasks.

\end{abstract}


%


\section{Introduction}
Motor learning, the acquisition and refinement of movement skills, is a fundamental aspect of human cognition and behaviour. Understanding its neural mechanisms is crucial for developing effective strategies in rehabilitation, sports training, and brain-computer interfaces. The use of EEG has been a longstanding approach in the investigation of neural activity underlying motor learning processes. EEG-based research has uncovered dynamic patterns of neural oscillations \citep{henz2016differential} and event-related synchronisation \citep{tan2014dynamic, haar2020brain} that correspond to the distinct learning mechanisms. These electrophysiological signatures have been associated with a variety of cognitive mechanisms essential to motor learning, such as attentional focus, sensorimotor integration, and memory consolidation \citep{Talsma2005Selective, Berti2008Cognitive}.

Reinforcement learning principles have provided insights into how the brain processes feedback and adapts behaviour based on previous experiences during motor learning. Neural activity patterns observed between trials have been found to reflect the processing of performance feedback from the previous trial, influencing subsequent motor planning and execution \citep{albert2016neural, luft2014learning, yuan2015cortical, hou2022gcns}. This feedback-dependent modulation of brain activity underscores the iterative nature of motor skill acquisition and the importance of trial-to-trial learning in shaping neural representations \citep{eliassen2012selective}.

\paragraph*{Related Work} Traditional statistical analysis approaches were often limited in fully capturing the spatial connections between different brain regions, which are essential for understanding the distributed nature of neural processing during complex tasks \citep{zaepffel2013modulations}. However, recent advancements in machine learning have driven the development of more sophisticated techniques. Neural networks and deep learning have emerged as powerful tools for analysing EEG data, enabling the automatic extraction of complex spatial-temporal representations from high-dimensional EEG signals \citep{schirrmeister2017deep, han2023noise, wang2018lstm}. Yet, these studies often face challenges in fully capturing the complex relationships between brain regions and ensuring the explainability of their findings.

Graph neural networks have emerged as a promising approach for analysing EEG data. These models can capture the temporal dynamics and spatial dependencies of neural activity, making them well-suited for studying the intricate network of brain connections involved in motor learning. However, the typical applications of deep learning have focused on static, well-structured brain data, such as that from motor imagery tasks \citep{hou2022gcns, han2023graph, sun2021adaptive}. Applying these powerful models to dynamic real-world brain data remains an area to be explored. This has the potential to enhance our understanding of how the brain learns and adapts, informing the development of more effective training protocols, rehabilitation strategies, and brain-computer interfaces. Furthermore, insights gained from studying brain patterns of complex motor skills could have broader implications for our understanding of cognitive processes, strategy, decision-making, and adaptation relevant to various aspects of human behavior.

In this work, we present a novel application of GNNs to investigate the neural correlates of motor learning and planning in a real-world task. Based on an EEG dataset that we collected during a real-world billiard task in an Embodied Virtual Reality environment \citep{nardi2024motor}, we trained a GNN model to predict the success or failure of the previous trial. The model leverages the EEG data from the pre-movement window, aiming to capture how the perceived success or failure of a given shot influences the neural processes involved in the motor planning of the next shot.
By applying GNNs to EEG data collected during a dynamic real-world motor task, unlike previous studies which focused on static brain imaging and motor imagery, we captured the intricate neural dynamics associated with motor planning, execution, and feedback processing in a more naturalistic setting \citep{tsay2024bridging}. Furthermore, we used multiple explainability methods and conducted extensive analysis to identify which brain regions and functional interactions were most salient in the model's prediction.
The results obtained contribute to our understanding of the neural mechanisms underlying motor skill acquisition and demonstrate the potential of GNNs as a powerful and explainable tool for the analysis of brain data.

\section{Methods}
\label{sec:methods}

\paragraph{Description of Dataset} The dataset utilised in our research originates from past experiments \citep{nardi2024motor}, where 40 healthy, inexperienced participants engaged in a real-world billiard task within our Embodied Virtual Reality (EVR) setup \citep{haar2020motor, haar2021embodied}. During the task, the participants performed a pool shot with the objective of pocketing a target ball, while immersed in a virtual environment that was accurately aligned with a physical setup, comprising a pool stick and a pool table. Throughout the task, brain activity was recorded with a wireless 14-channel EEG headset (Emotiv EPOC+) with a 256Hz sampling rate. The outcome variable examined was the \textit{success} or \textit{failure} during the previous trial, as we aimed to investigate any changes in brain activity resulting from the previous feedback. 

The 2-second window preceding each shot was the focus of our analysis. This timeframe was selected to capture motor planning processes, occurring before the ``main" movement, and to exclude any visual feedback from the model predictions, as participants observed the ball trajectory only after the shot was performed. Indeed, this window was chosen to avoid including signals unrelated to the primary movement, thereby directing our GNN model training to the most relevant data.

Participants took part in two learning rounds with different feedback conditions and shooting towards different pockets. Here we analyse brain activity only from the round with the error feedback task \citep{nardi2024motor}, from four groups of 10 participants, differing by round and pocket: \textit{First Left} (first round, shooting at the left pocket), \textit{First Right} (first round, shooting at the right pocket), \textit{Second Left} (second round, shooting towards the left pocket), and \textit{Second Right} (second round, shooting towards the right pocket).

\paragraph{EEG Data Pre-processing} EEG data was band-pass filtered at 1-45Hz using FIR filter, and cleaned by removing artifacts from the signals, including bad data segments and problematic channels, employing the \textit{Artifact Subspace Reconstruction} algorithm \citep{kothe2013artifact}. Furthermore, signal windows were removed based on thresholds for the maximum standard deviation of bursts and the maximum fraction of contaminated channels allowed in the final output data for each considered window. The prefrontal channels (AF3, AF4) were removed from the EEG data to avoid the models wrongly classifying changes in eye movement patterns instead of the proper underlying brain signal, resulting in a 12-channel dataset. The data was segmented into 2-second epochs preceding each shot, as the motor planning window. An independent component analysis (ICA) was conducted on the data to eliminate muscle, eye, and cardiac artefacts. The missing channels were subsequently interpolated from the remaining ones using spherical spline interpolation and a topographical map. This preprocessing was informed by prior research on motor learning \citep[e.g.][]{tan2014dynamic, haar2020brain} and based on the 10-20 international EEG system.

Due to the limited data available for each participant, we combined together the 10 participants within each group. We created a balanced dataset for each group by accounting for the number of successful and unsuccessful trials, to ensure that the models could predict both classes with equal proficiency. To do so, we examined the angular error of each shot relative to the pocket, and removed the \textit{failure} trials closest to the success region, or the worst successful trials - farthest from the centre of the pocket. This selection was based on the expectation that any brain signals related to success feedback would be more prominent for shots greatly differing from the opposite class.
To ensure sufficient distinction between trials belonging to different classes, we considered failures that were at least 2 degrees off from the success area. Additionally, we removed the first trial of each block, as the preceding trial was not performed immediately before it. One participant was excluded from the study due to the too low number of trials after the EEG preprocessing pipeline.

\paragraph{Graph Formation and Model Architecture} To model EEG data as brain graphs, we constructed an adjacency matrix $\mathbf{A}$, where $A_{i,j}$ is each term of the matrix that indicates the connection strength between EEG channels $i$ and $j$. This strength is computed based on spatial proximity, with closer channels having stronger connections. We used a ball tree algorithm to assign channels within a predefined radius as neighbours and calculate the edge weights inversely proportional to the squared distance between channel coordinates, setting the maximum distance between neighbours to 75\% of the head radius.

Given the relatively small size of our dataset and the limited number of channels, we designed a shallow temporal CNN and GNN architecture to mitigate the risk of overfitting (see Figure \ref{fig:ModelStructure}). The model processes raw EEG signals with a 1D convolution to capture temporal features, followed by batch normalisation, Parametric ReLU activation, average pooling, and dropout. These features are then input into a Chebyshev graph convolution layer \citep{defferrard2016convolutional}, which propagates information across the graph based on the adjacency matrix $\mathbf{A}$. The output is globally averaged and passed through a linear classification layer. The model is trained with a cross-entropy loss function, using class weights, balancing accurate classification with regularisation. The detailed methodology for constructing the graph representation and the complete neural network architecture are provided in Appendix \ref{apd:graph}.

\begin{figure}[t]
    \floatconts
    {fig:ModelStructure}
    {\caption{An Illustration of our Graph Neural Network Architecture.}}
    {\includegraphics[width=\linewidth]{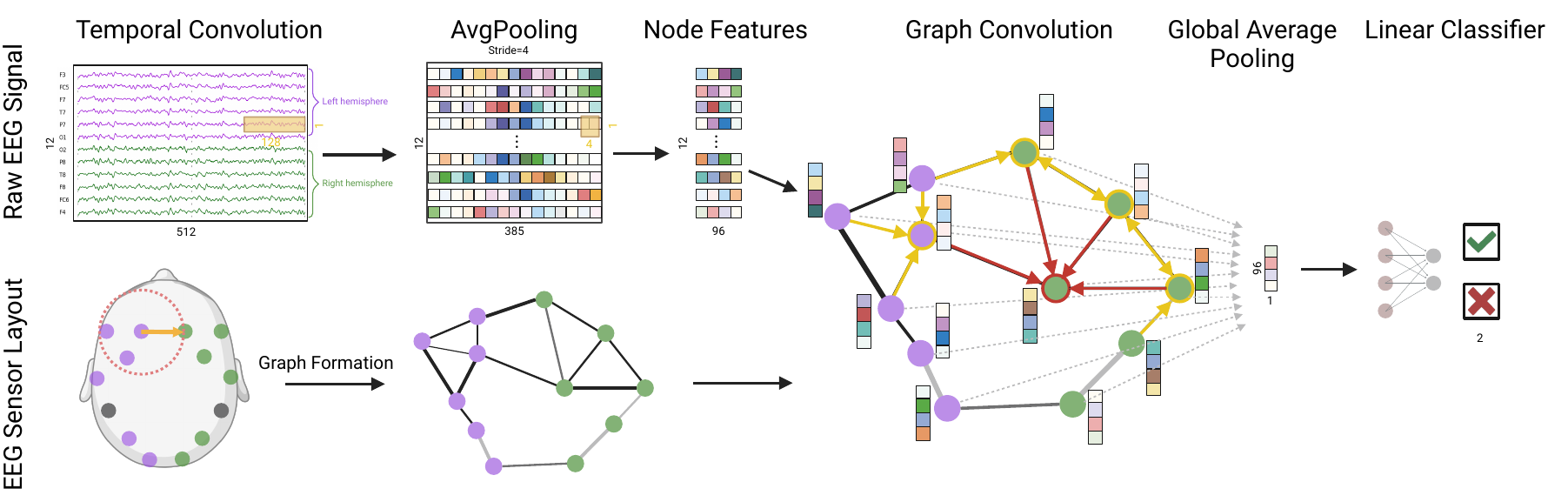}}
\end{figure}

\paragraph{GNN Spatial Explainability} To investigate the differences between the models applied to the four groups, we implemented our previous explainability pipeline that seeks to identify the most salient channels contributing to the model predictions \citep{han2023graph}. We implemented node and edge masking techniques based on the \textit{GNNExplainer} method \citep{ying2019gnnexplainer} to investigate the spatial characteristics of our biomarker. This approach involved freezing the weights of the trained GNN model and applying learnable node and edge masks to the raw data and the GNN's edges, respectively. The sigmoid function was utilised to encourage binary masking, enhancing the model's ability to highlight significant spatial features during the biomarker identification process.

The learnable masks were trained using a loss function (\ref{eq:Loss}) which combines classification edge mask size loss $L_{EMS}$, node mask entropy loss $L_{NME}$, edge mask entropy loss $L_{EME}$, cross-entropy loss $L_{CES}$, and node mask size loss $L_{NMS}$. 

\begin{align}\label{eq:Loss}
    L_{total} = L_{EMS} + L_{NME} + L_{EME} + L_{CES} + L_{NMS}
\end{align}

 The classification loss ensured the masked model maintained the original classification accuracy, while the size and entropy losses promoted the masks to be sparse and discrete. The importance score of a node or edge was determined by the corresponding mask value after optimisation, as higher mask values indicated higher importance for the model to minimise the classification loss. 

The purpose of this explainability technique is to identify the most salient EEG channels that contribute significantly to the model's predictions. By applying learnable node and edge masks to the raw data and the graph neural network's edges, we can determine the importance of each spatial feature in the classification process. The goal is to find the minimal set of nodes that can be removed while maintaining comparable classification accuracy, thus highlighting the most influential channels for the model's decisions.

\paragraph{Distance Measures between Maps} To compare the similarity or divergence between the contribution maps, we utilised a modified version of the Gromov-Wasserstein distance, known as the Sliced Gromov-Wasserstein (SGW) distance, which is less intensive computationally than the Gromov-Wasserstein distance, while preserving the key advantages of the original metric \citep{vayer2019sliced}. 
The SGW distance between two graphs is defined as

\begin{align}\label{eq:SGW}
d_{SGW}(G_1, G_2) &= \mathbb{E}_{\theta \sim \mathcal{U}(S^{d-1})} \left[ W_1\left(p_{\theta}(G_1), p_{\theta}(G_2)\right) \right] 
\end{align}
where $G_1$ and $G_2$ are the two graphs being compared, $p_\theta(G)$ represents the projection of graph $G$ onto the random direction $\theta$, which is sampled uniformly from the unit sphere $S^{d-1}$. $W_1(p_\theta(G_1), p_\theta(G_2))$ is the 1D-Wasserstein distance between the projected distributions $p_\theta(G_1)$ and $p_\theta(G_2)$. The final SGW distance is the expected value of these 1D-Wasserstein distances over all the random directions $\theta$.

To define the maps for comparison, we transformed the Cartesian electrode coordinates into polar coordinates and added the mask value as the third dimension. This approach allowed us to consider both the mask differences and the spatial information when comparing the maps. To assess the statistical significance of the differences observed between the contribution maps across the various conditions, we performed Mann-Whitney U tests. This non-parametric test was chosen as it does not require the data to follow a normal distribution. In case of paired data, when the samples derived from the same underlying distribution, we performed paired Wilcoxon signed-rank tests.

\paragraph{Training and Evaluation} To improve model performance, and based on the assumption of shared underlying features across the groups, we implemented a pre-training strategy prior to the main model training, technique shown to be effective in boosting performances \citep{hu2019strategies}. Although pretraining is typically defined as training the model on a separate task \citep{erhan2010does}, the uniqueness of the experimental data led us to instead pretrain each group on the other groups. Specifically, we implemented two pretraining approaches: one based on the \textit{Pocket}, and the other on the \textit{Round}. That is, we pretrained each group for 200 epochs on the two other groups belonging to the opposite pocket side or the opposite round, in order to have two comparable groups in every pretraining scenario. This pretraining approach allowed us to leverage the shared features across the groups, potentially enhancing the model's ability to generalise and improve its performance.

Following pretraining, a 10-fold cross-validation training procedure was then performed. The split in folds was done ensuring that each participant was evenly represented across the folds, while also maintaining a balance between the classes, due to the individual variability in brain data. In this case, the model was trained for only 150 epochs due to the smaller sample size. Both models were trained using a \texttt{batch size} of 32 and a \texttt{learning rate} of 1e-4 to ensure a trade-off between optimality and training costs. To prevent overfitting, we implemented L2-optimisation in the \texttt{Adam} optimiser, edge dropout in the initial graph ($p=0.2$), and dropout at the end of the temporal convolution layer ($p=0.35$) in all the models \citep{zhang2018improved, rong2019dropedge, srivastava2014dropout}. To evaluate the models performances, we examined accuracy, precision, recall, and F1-score on the validation set, which was not used for any training decisions or hyperparameter optimisation, to assess the models' ability to generalise and perform on unseen data.

\section{Results}

\paragraph{Model Performance} The pretrained models, based on the \textit{Round} or \textit{Pocket} groupings, led to significant improvements in model performance across all evaluation metrics. Without pretraining, our GNN models struggled to extract meaningful patterns from the EEG data, achieving an average accuracy of only 0.54 across all groups, barely above chance level. The introduction of pretraining, however, led to a significant boost in model performance. Both \textit{Round}-based (Figure \ref{fig:PerformanceMetrics}B) and \textit{Pocket}-based (Figure \ref{fig:PerformanceMetrics}C) pretraining resulted in substantial improvements, with average accuracies across groups increasing to 65\% and 66\% respectively, with values up to 70\% in the individual groups. This consistent enhancement across all participant groups suggests that pretraining enables the GNNs to learn more robust and generalisable features from the EEG data. The full metrics comparison is shown in Appendix \ref{apd:modelPerformances}. 

\begin{figure}[bt]
\floatconts
  {fig:PerformanceMetrics}
  {\caption{Performances of GNN models with (\textbf{A}) no pretraining, (\textbf{B}) pretraining on opposite \textit{Round}, (\textbf{C}) and pretraining on opposite \textit{Pocket}. The 10-fold cross validation accuracy distributions of the four groups are displayed separately for each pretraining mode. Black dashed line represents chance level, while the green dashed line shows the average accuracy across groups for the individual mode.}
  }
  {\includegraphics[width=\linewidth]{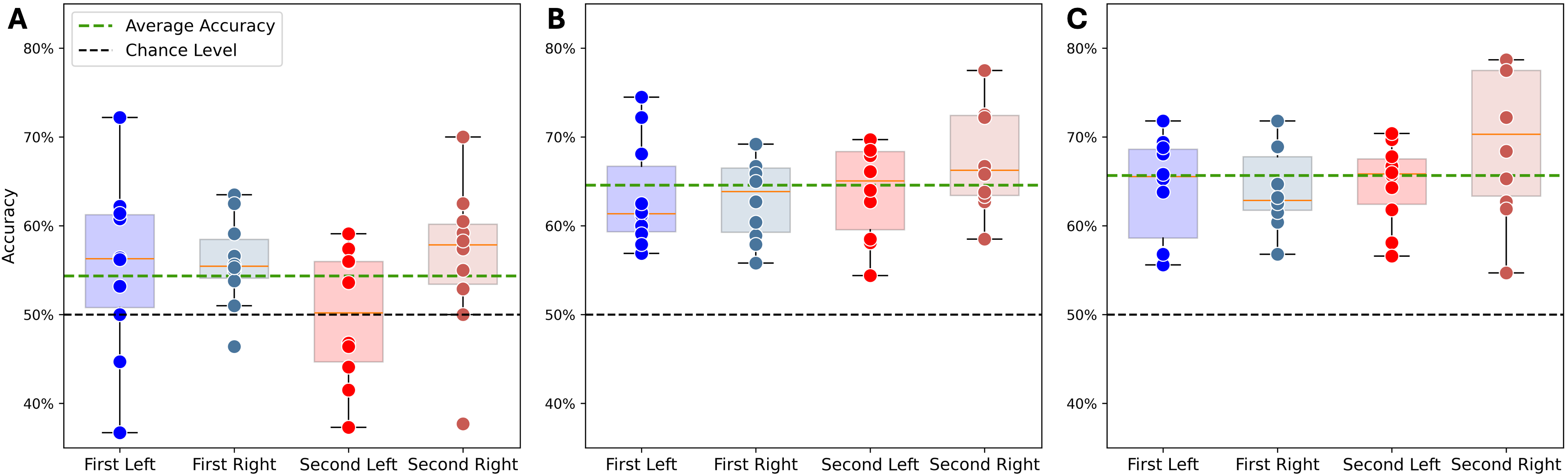}}
\end{figure}

\begin{figure}[bt]
\floatconts
  {fig:BrainMaps}
  {\caption{Brain activation maps across the four groups (\textit{First Left, First Right, Second Left, Second Right}). Arrows connect the pretraining maps used, based on opposite round (first column) or opposite pocket (fourth column) to the final maps of the four groups.}}
  {\includegraphics[width=0.97\linewidth]{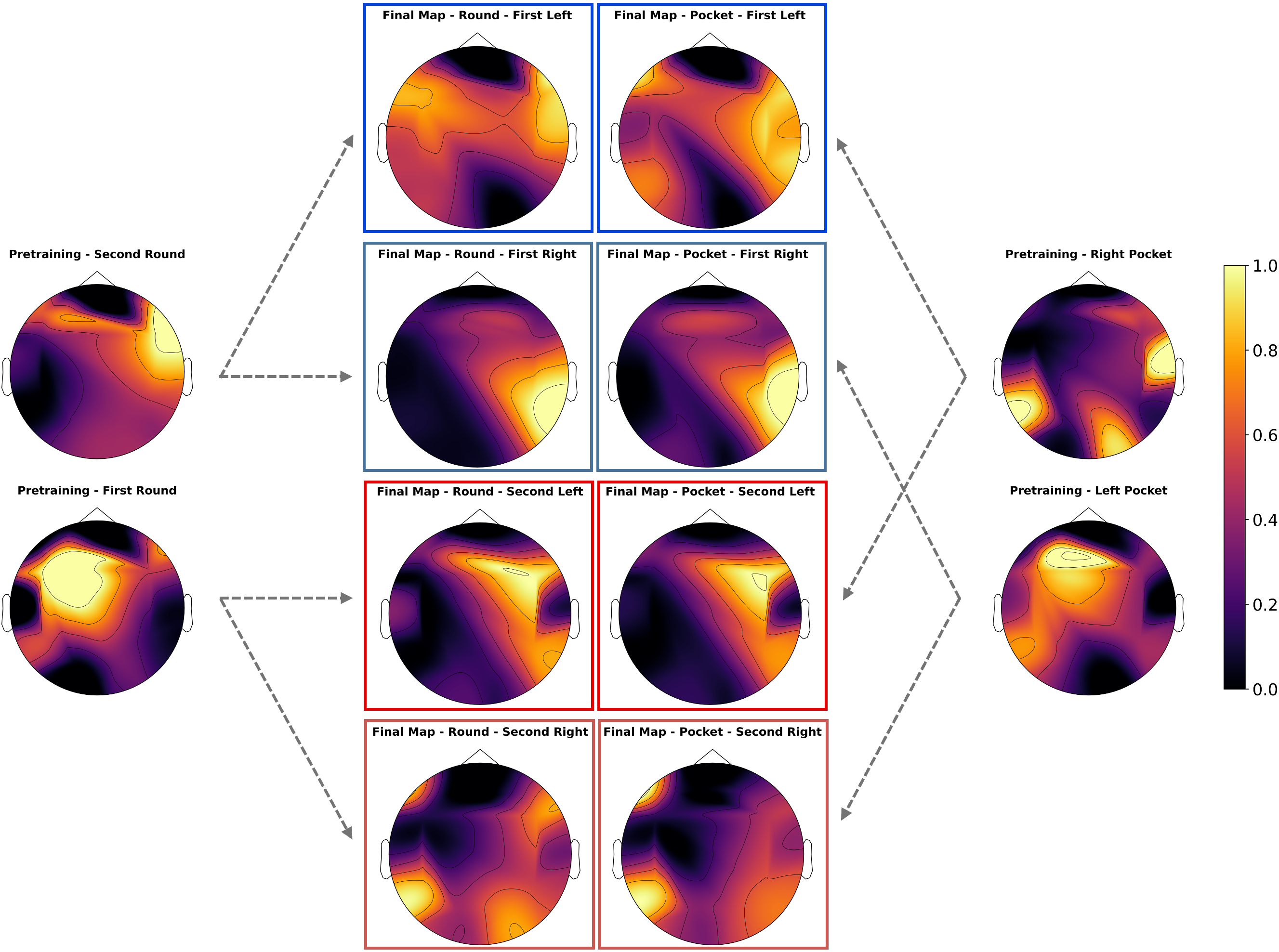}}
\end{figure}

\paragraph{Explainability Results} The explainability analysis of our GNN models revealed consistent group-specific neural signatures across different pretraining conditions. While the \textit{Round} or \textit{Pocket} pretraining maps are very different, the brain maps for each group exhibited similar spatial patterns of importance, irrespective of the pretraining method (Figure \ref{fig:BrainMaps}). This consistency suggests that the models have identified stable neural representations for each group. The persistence of these group-specific patterns across pretraining conditions indicates the presence of underlying neural architectures unique to each group's learning experience. This finding implies potential differences in the effect of reinforcement learning from the previous trial based on aspects of the task.
To quantify the similarities between these neural representations, we measured the Sliced Gromov-Wasserstein (SGW) distances (Figure \ref{fig:RSM}A), which showed the highest similarities between the four groups to themselves across the different pretraining conditions. This structure provides evidence for the stability of group-specific neural patterns, suggesting that the characteristics of each group's brain activity during the task remain relatively consistent and robust to varying training conditions of the models.

Comparing the distances of each group's pretraining map to their final map (Figure \ref{fig:RSM}B - II-III) and the distances between the final maps of the groups (Figure \ref{fig:RSM}B - IV), it is evident that the GNNs primarily relied on the group-specific spatial information rather than the underlying common information. Similarly, the significant difference ($p=0.03$) between the distances of groups starting from the same pretraining (Figure \ref{fig:RSM}B - V) and the distances between the final maps (Figure \ref{fig:RSM}B - IV) represents the distinctiveness of each group's neural representations, suggesting the minor importance of the pretraining strategy. Noticeable distributional differences were found between the remaining features, although these differences were not statistically significant due to the small sample size.

These results suggest that despite the complex nature of brain activity during motor tasks, there exist discernible, group-specific patterns that can be identified by machine learning models. The ability of the models to capture these consistent patterns across different pretraining strategies demonstrates the adaptability of the GNN architecture in processing complex EEG data, suggesting that these models can extract relevant features from the spatial relationships between EEG channels, providing insights into the neural basis of motor learning and feedback processing.

\begin{figure}[bt]
\floatconts
  {fig:RSM}
  {\caption{\textbf{(A)} Representation Distance Matrix between group maps \textbf{(B)} Distribution of distances between groups of: (I) same pretraining strategies; final vs initial maps when pretraining with opposite (II) round and (III) pocket; (IV) final maps within group with different pretraining; final maps between groups that had the same pretraining maps as starting point (V). The quantities reported in the boxplots are framed in the RDM using the same colour scheme.}}
  {\includegraphics[width=\linewidth]{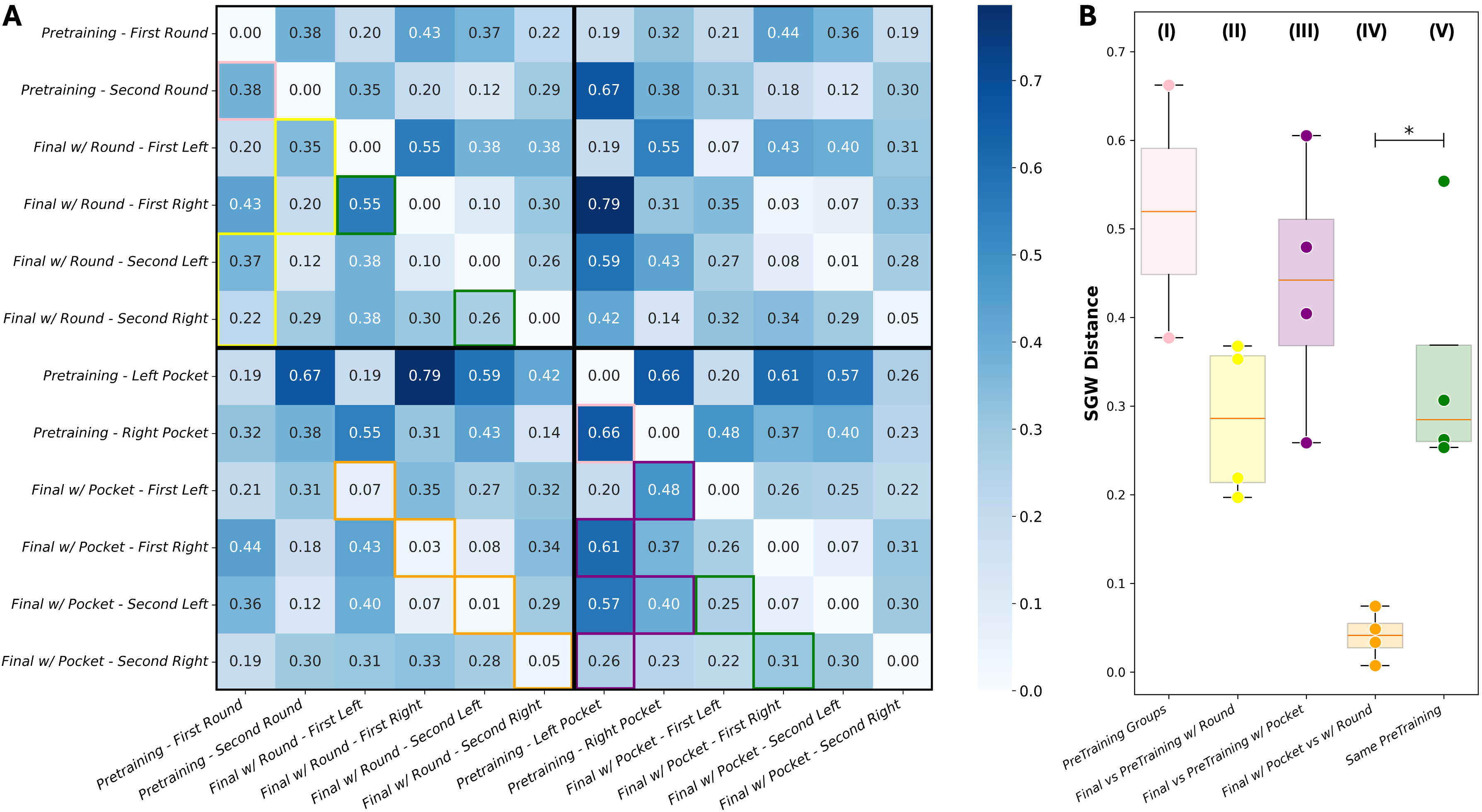}}
\end{figure}

\section{Discussion}

This study used GNNs to understand the neural correlates of performance feedback in motor planning in a real-world billiard task. By analysing the spatial relationships in the EEG data, our models revealed consistent group-specific neural signatures, providing insights into the complex brain activity underlying motor skill acquisition. 
GNNs have shown great promise in analysing static, well-structured brain data, such as that from motor imagery tasks \citep{hou2022gcns, sun2022graph}. Previous research has highlighted the ability of GNNs to effectively capture spatial relationships in brain data \citep{kim2021learning, azevedo2022deep}. Yet, the successful application of these models to dynamic, real-world tasks remains an active area of research, where novel techniques are needed to overcome the inherent complexities of brain activity in natural environments.

Here we manipulated the GNNs training pipeline, implementing different pretraining strategies using both opposite \textit{Round} or \textit{Pocket}, which led to a significant improvement in model performance, despite the unique and complex nature of the task, indicating that the pretraining process enabled the GNNs to acquire more robust and generalisable features from the EEG data. This consistent enhancement suggests that the pretraining methods allowed the models to better navigate the intricate manifold of neural representations, coherently with previous studies \citep{hu2019strategies, hu2020gpt}. Notably, the comparable performance gains for both pretraining approaches imply that the GNNs were able to effectively leverage different aspects of the experimental design to enhance their predictive capabilities, hinting at the presence of underlying neural patterns that persist across experimental conditions.

Interestingly, persistent group-specific patterns were observed in the brain maps generated through our explainability analysis. The GNN models exhibited robust spatial invariance, and consistently identified distinct spatial activation patterns for each of the four groups, regardless of the pretraining approach. This consistency of group-specific neural representations suggests that individual differences in motor learning strategies or feedback processing mechanisms are reflected in stable, identifiable patterns of brain activity.

Taken together, these results suggest that the neural correlates underlying motor learning and performance feedback processing are influenced by both individual differences in learning strategies as well as the specific demands and characteristics of the task. The consistent detection of group-specific neural representations by the GNN models, irrespective of the pretraining method, underscores the robustness of our approach in extracting meaningful neural signatures associated with complex cognitive and motor processes.

\paragraph{Limitations and Future Work} This study provides valuable insights into the neural correlates of motor learning using GNNs, but several limitations exist. The small sample size restricts the generalisability of the findings and may have contributed to sub-optimal model performance, suggesting the need to increase the sample size in future studies which could enhance accuracy and robustness. Individual differences in brain activity and learning strategies also remain a challenge, suggesting the need for more personalised modelling approaches and the integration of additional transfer learning techniques. 
Further work is needed to connect the spatial patterns in EEG to specific cognitive processes and neurophysiological mechanisms, as well as investigate the temporal aspects of motor planning. Lastly, the study focused on short-term feedback, leaving open the question of how neural representations change over longer periods of skill acquisition.

\paragraph{Conclusion} This study highlights the potential of Graph Neural Networks to provide meaningful insights into complex real-world neural data. The consistent group-specific spatial patterns found underscore the robustness of this method in capturing differences in the underlying cognitive and motor processes. This study lays the groundwork for future advancements in understanding complex neural representations using GNNs.

\section*{Acknowledgements}

F.N. is supported by UK Research and Innovation [UKRI Centre for Doctoral Training in AI for Healthcare grant number EP/S023283/1]; S.H. is supported by the Edmond and Lily Safra Fellowship Program and by the UK Dementia Research Institute Care Research \& Technology Centre; A.A.F. acknowledges a UKRI Turing AI Fellowship Grant (EP/V025449/1).

\newpage

\bibliography{pmlr-sample}

\appendix
\newpage
\section{Experimental Setup and Design}

\subsection{EVR Setup}

\begin{figure}[ph]
\floatconts
  {fig:image}
  {\caption{(\textbf{A}) Billiard task in the Embodied Virtual Reality setup: participants playing pool in real-world (\textit{right}) are immersed in the VR environment in which visual feedback manipulations are applied (\textit{left}); (\textbf{B}) 12-channels EEG data sample used for the GNN models: out of a window of 4 seconds around each trial, only the 2s pre-shot window is considered for the analysis; (\textbf{C}) Experimental conditions of the four groups: combinations of \textit{Left} and \textit{Right} pockets and \textit{First} and \textit{Second} rounds were considered according to the session and shooting direction experienced by the participants.}}
  {\includegraphics[width=\linewidth]{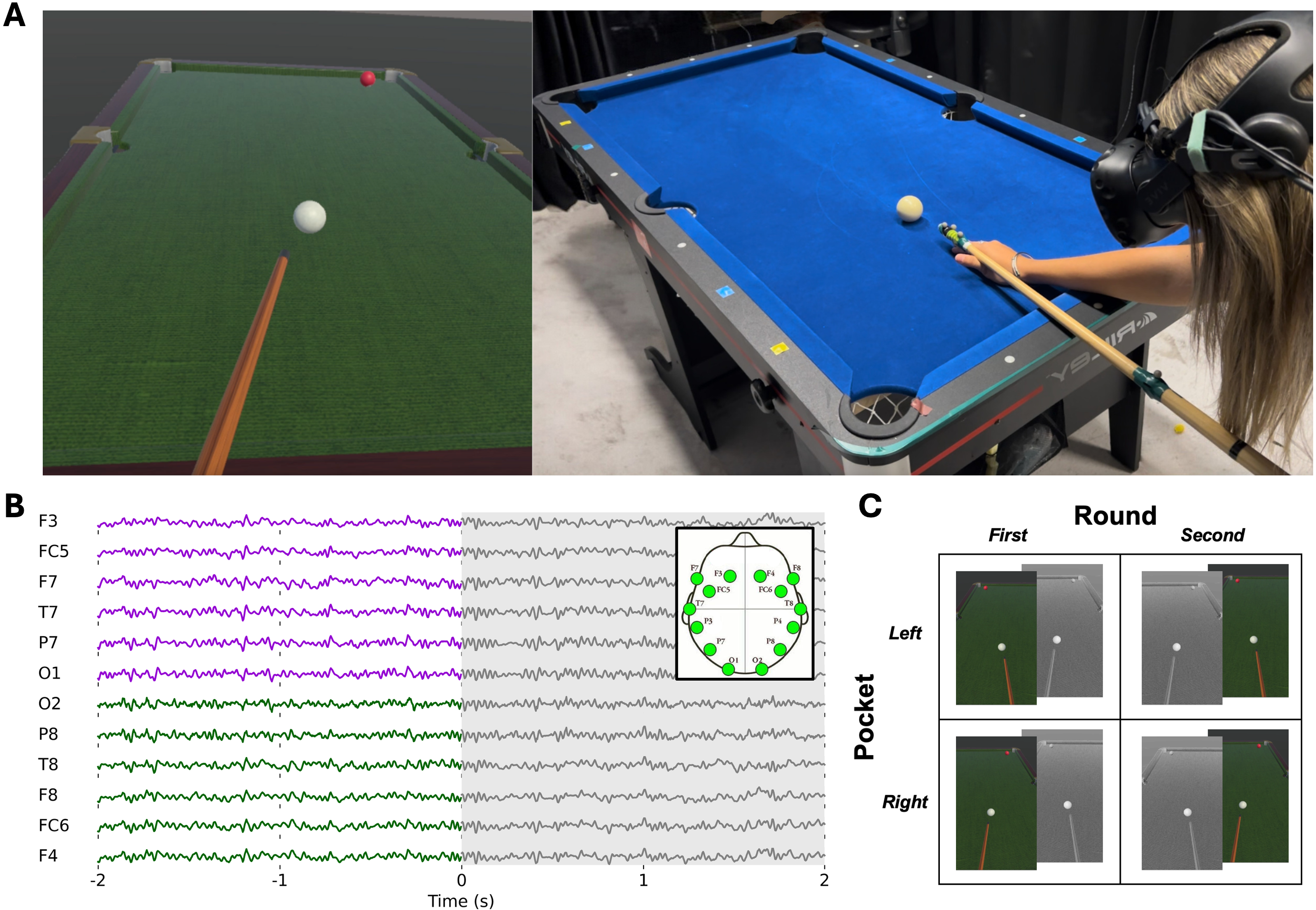}}
\end{figure}

\newpage

\subsection{Label Derivation}

\begin{figure}[h]
\floatconts
  {fig:image}
  {\caption{Distribution of all trials of all participants over the standardised angle range (0° represents a perfect successful shot), divided by label. The \textit{No Decision} category includes the failure trials which are too close to the success funnel (in green), according to a pre-determined threshold of 2 degrees. $N=39$.}}
  {\includegraphics[width=0.80\linewidth]{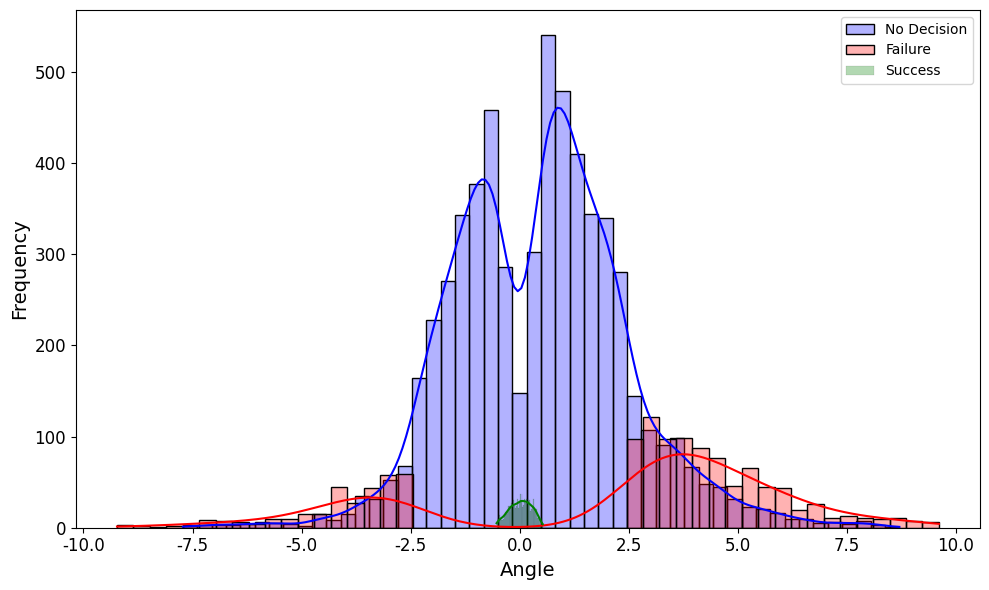}}
\end{figure}

\newpage
\section{Graph Formation and Model Architecture}\label{apd:graph}

\subsection{Graph Formation}
To represent our EEG data as brain graphs, we first construct an adjacency matrix $\mathbf{A}$, where $\mathbf{A} \in \mathbb{R}^{C \times C}$, with $C$ denoting the total number of EEG channels (nodes). The elements of the matrix, $A_{i,j}$, represent the connection strength or edge weight between nodes $i$ and $j$, which in our case correspond to the spatial proximity between EEG channels. A stronger connection between two nodes results in a higher edge weight, while an absence of connection is indicated by $A_{i,j}=0$.

Given the spatial coordinates of the EEG channels, we employ a ball tree algorithm, called the \texttt{cKDTree}, to efficiently identify neighbouring channels within a predefined radius $r$ \citep{narasimhulu2021ckd}. For each channel $i$, the algorithm identifies all channels $j$ within this radius, forming potential edges between them. We empirically found that $r=75\%$ head radius was the best trade-off between a low number of neighbours for each node, to avoid misleading information coming from very far channels, and the inclusion of all the channels in the headset.

The weight of the edge between nodes $i$ and $j$ is then computed as the inverse square of the Euclidean distance between their corresponding coordinates. Mathematically, the weight $w_{i,j}$ of the edge is given by:

\begin{equation}
w_{i,j} = 
\begin{cases} 
\frac{1}{\|\mathbf{x}_i - \mathbf{x}_j\|^2} & \text{if } \|\mathbf{x}_i - \mathbf{x}_j\| \neq 0, \\
0 & \text{if } \|\mathbf{x}_i - \mathbf{x}_j\| = 0,
\end{cases}
\end{equation}

where $\mathbf{x}_i$ and $\mathbf{x}_j$ are the spatial coordinates of channels $i$ and $j$, respectively, and $\|\mathbf{x}_i - \mathbf{x}_j\|$ denotes the Euclidean distance between these channels. The inverse square relationship emphasizes stronger connections between closer nodes while diminishing the influence of more distant nodes. Self-links (i.e., $A_{i,i}$) are avoided to ensure the adjacency matrix strictly represents inter-channel relationships.

The resulting adjacency matrix $\mathbf{A}$ thus captures the spatial structure of the EEG channels, supported by neurophysiological findings that suggest connective strength diminishes with squared distance. 

\subsection{GNN Architecture}
We designed the model to have a shallow temporal CNN and GNN architecture, in order to mitigate the risk of overfitting due to the relatively small sample size. Our model takes in two primary inputs: the raw EEG signals, represented as node features with dimensions $C \times T$, where $C$ denotes the number of channels (nodes) and $T$ signifies the number of time points, and the structural information of the graph, represented by the $C \times C$ adjacency matrix $\mathbf{A}$, which encodes the connectivity and edge weights among the nodes.

The architecture begins by applying a one-dimensional convolution (1D-CNN) along the temporal axis of the raw EEG signals. The convolution operation uses a kernel size of $\frac{f_s}{2}$, where $f_s$ is the sampling rate of the EEG signals \citep{lawhern2018eegnet}. This kernel size is chosen to capture the relevant frequency information in the EEG signals effectively. The temporal convolution layer is defined as:

\begin{equation}
\mathbf{H}_1 = \text{PReLU}(\mathbf{W}_{\text{conv}} * \mathbf{X} + \mathbf{b}_{\text{conv}}),
\end{equation}

where $\mathbf{X} \in \mathbb{R}^{C \times T}$ is the input EEG data, $\mathbf{W}_{\text{conv}}$ is the convolutional kernel, $\mathbf{b}_{\text{conv}}$ is the bias term, and $*$ denotes the convolution operation along the time axis. After the convolution, we apply batch normalisation and a Parametric ReLU (PReLU) activation function to introduce non-linearity, followed by average pooling and dropout layers. The average pooling layer reduces the temporal dimensionality, thereby lowering the risk of overfitting in subsequent layers.

Following temporal feature extraction, the reduced node features are fed into a single Chebyshev graph convolution layer to propagate information across neighbouring nodes (EEG channels). The Chebyshev graph convolution is defined as:
\begin{equation}
\mathbf{H}_2 = \sum_{k=0}^{K-1} \mathbf{T}_k(\tilde{\mathbf{L}}) \mathbf{H}_1 \mathbf{\Theta}_k,
\end{equation}
where $\tilde{\mathbf{L}} = 2\mathbf{L}/\lambda_{\max} - \mathbf{I}_C$ is the rescaled Laplacian matrix derived from the adjacency matrix $\mathbf{A}$, $\mathbf{L} = \mathbf{D} - \mathbf{A}$ is the graph Laplacian, $\mathbf{D}$ is the degree matrix, $\mathbf{T}_k(\cdot)$ represents the Chebyshev polynomial of order $k$, $\lambda_{\max}$ is the maximum eigenvalue of $\mathbf{L}$, and $\mathbf{\Theta}_k$ are the learnable parameters of the layer. This formulation allows for localised spectral filtering, enabling the model to capture complex spatial dependencies among EEG channels while maintaining computational efficiency.

The output from the Chebyshev GNN layer, $\mathbf{H}_2$, undergoes a \textit{Softplus} activation to ensure non-negative activations. Subsequently, global average pooling is applied across all node features to aggregate the information into a single feature vector. Finally, the aggregated feature vector $\mathbf{h}_{\text{global}}$ is passed through a linear classification layer to perform the final prediction. Overall, all  layers in the architecture are trained to minimise a class-weighted cross-entropy loss function, defined as:

\begin{equation}
\mathcal{L} = -\frac{1}{N} \sum_{n=1}^{N} \sum_{c=1}^{C} y_{n,c} \log(\hat{y}_{n,c}),
\end{equation}

where $N$ is the number of samples, $C$ is the number of classes, $y_{n,c}$ is the true label of class $c$ for sample $n$, and $\hat{y}_{n,c}$ is the predicted probability for class $c$ after applying a softmax function to the output of the linear classification layer.

This simple architecture efficiently leverages both temporal and spatial information from the EEG data, allowing for accurate classification while minimising the risk of overfitting.

\newpage
\section{Model Performances}\label{apd:modelPerformances}

\begin{table}[bthp]
\floatconts
  {tab:ModelPerformances}
  {\scriptsize
  \resizebox{\textwidth}{!}{
  \begin{tabular}{lcccc|cccc|cccc} 
    \toprule
    & \multicolumn{4}{c|}{\textbf{No Pretraining}} & \multicolumn{4}{c|}{\textbf{Pretraining w/ Round}} & \multicolumn{4}{c}{\textbf{Pretraining w/ Pocket}} \\ 
    \cline{2-13} & \textit{Acc} & \textit{Prec} & \textit{Rec} & \textit{F1} & \textit{Acc} & \textit{Prec} & \textit{Rec} & \textit{F1} & \textit{Acc} & \textit{Prec} & \textit{Rec} & \textit{F1}  \\ 
    \cline{2-13}
    \textit{First Left} & 0.55 & 0.55 & 0.57 & 0.55 & 0.63 & 0.61 & 0.74  & 0.67 & 0.64  & 0.61   & 0.80  & 0.69  \\
    \textit{First Right}     & 0.56  & 0.53   & 0.57  & 0.54 & 0.63  & 0.63   & 0.65  & 0.64 & 0.64  & 0.64   & 0.66  & 0.65  \\
    \textit{Second Left}     & 0.50  & 0.50   & 0.42  & 0.44 & 0.64  & 0.67   & 0.55  & 0.60 & 0.65  & 0.64   & 0.67  & 0.66  \\
    \textit{Second Right} & 0.56  & 0.58   & 0.53  & 0.54 & 0.68  & 0.69   & 0.65  & 0.67 & 0.70  & 0.71   & 0.68  & 0.69  \\
    \cline{1-13} 
    \textit{Average} &  0.54 & 0.54 & 0.52 & 0.52 & 0.65 & 0.65 & 0.65 & 0.65 & 0.66 & 0.65 & 0.70 & 0.67 \\
    \bottomrule
  \end{tabular}
  }}
  {\caption{GNN Model Performances for the four groups, obtained as the average of the performances on the validation sets of the 10-fold cross validation. The three groups of columns show metrics without any pretraining, pretraining on opposite \textit{Round} and opposite \textit{Pocket}, respectively. The last row shows the average for the metrics across the four groups. \\
  \textit{Acc}: Accuracy; \textit{Prec}: Precision; \textit{Rec}: Recall; \textit{F1}: F1-score.}}
\end{table}

\newpage
\section{GNN Explainability Maps - Evolution over Learning}\label{apd:MapsEvolution}

\begin{figure}[h]
\floatconts
  {fig:image}
  {\caption{Brain Maps Evolution across learning when pretraining on opposite \textit{Round} (A-D) and \textit{Pocket} (E-H), for the \textit{First Left} (A,E), \textit{First Right} (B,F), \textit{Second Left} (C,G), and \textit{Second Right} (D, H) groups. The five columns show the explainability maps for pretraining, epochs 10, 20, 50 and 150 respectively. The colour bar refers to all the plots and represents the standardised contribution of the different channels, according to explainability methods described in Section \ref{sec:methods}. \vspace{-0.5cm}}}
  {\includegraphics[width=0.558\linewidth]{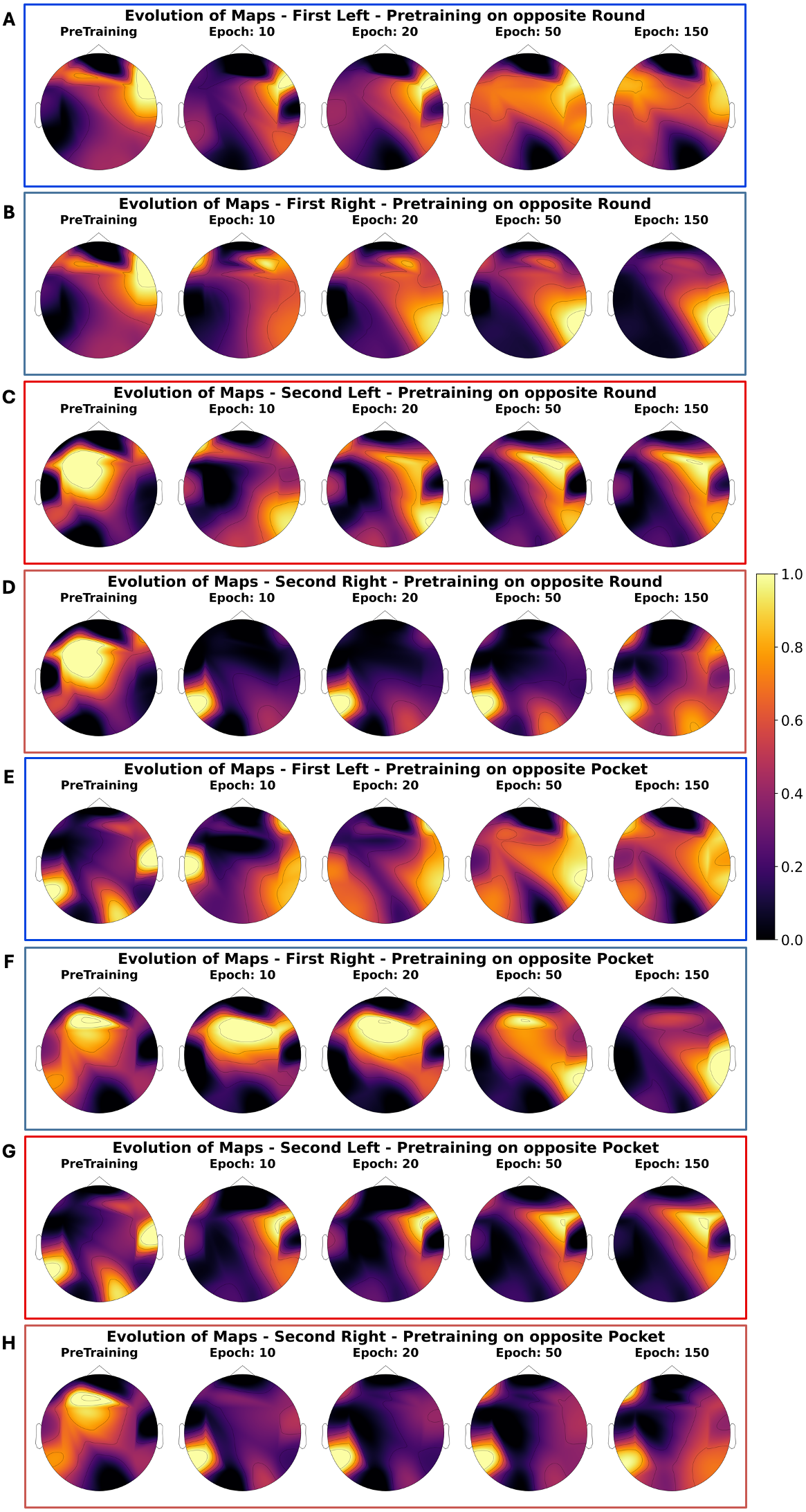}}
\end{figure}
\end{document}